\documentclass[11pt,journal]{IEEEtran}

\ifCLASSOPTIONcompsoc
\usepackage[nocompress]{cite}
\else
\usepackage{cite}
\fi
\usepackage{cite}
\usepackage{hyperref}
\usepackage{url}

\usepackage{amsmath,graphicx,amssymb,amstext,fixltx2e}
\usepackage{verbatim}
\usepackage{subfig}
\usepackage{enumerate}
\usepackage{cite}
\usepackage{float, subfloat}
\usepackage{tabularx}
\usepackage{multirow}
\usepackage{bigstrut}
\usepackage{amsmath}
\usepackage[table,xcdraw]{xcolor}
\usepackage{makecell}
\usepackage{amsfonts}

\DeclareMathAlphabet{\mathsfsl}{OT1}{cmss}{m}{sl}

\begin{document}

	\title{The feasibility of automated identification of six algae types using neural networks and fluorescence-based spectral-morphological features}

	\author{Jason L. Deglint, Chao Jin, Angela Chao, Alexander~Wong,~\IEEEmembership{Senior Member,~IEEE}
		\IEEEcompsocitemizethanks{
			\IEEEcompsocthanksitem J.L.~Deglint, C.~Jin, A~Chao, and A.~Wong are with the Department
			of  Systems Design Engineering, University of Waterloo, Waterloo,
			ON, Canada.  \protect\\
			E-mails: jdeglint@uwaterloo.ca, j3chao@uwaterloo.ca
		}
		\thanks{
			
			The authors thank the Natural Sciences and Engineering Research Council of Canada and the Canada Research Chairs Program.
			
	}}

	\markboth{IEEE}%
	{Deglint \MakeLowercase{\textit{et al.}}: The feasibility of automated identification of six algae types using neural networks and fluorescence-based spectral-morphological features}
	
	\maketitle

\begin{abstract}
Harmful algae blooms (HABs), which produce lethal toxins, are a growing global concern since they negatively affect the quality of drinking water and have major negative impact on wildlife, the fishing industry, as well as tourism and recreational water use. The gold-standard process employed in the field to identify and enumerate algae requires highly trained professionals to manually observe algae under a microscope, which is a very time-consuming and tedious task.  Therefore, an automated approach to identify and enumerate these micro-organisms is much needed.  In this study, we investigate the feasibility of leveraging machine learning and fluorescence-based spectral-morphological features to enable the identification of six different algae types in an automated fashion.  More specifically, a custom multi-band fluorescence imaging microscope is used to capture fluorescence imaging data of a water sample at six different excitation wavelengths ranging from 405 nm - 530 nm.  Automated data processing and segmentation was performed on the captured fluorescence imaging data to isolate different micro-organisms from the water sample.  A number of morphological and spectral fluorescence features are then extracted from the isolated micro-organism imaging data, and used to train neural network classification models designed for the purpose of identification of the six algae types given an isolated micro-organism.  Experimental results using three different neural network classification models (one trained on morphological features, one trained on fluorescence-based spectral features, and one trained on fluorescence-based spectral-morphological features) showed that the use of either fluorescence-based spectral features or fluorescence-based spectral-morphological features to train neural network classification models led to statistically significant improvements in identification accuracy when compared to the use of morphological features (with average identification accuracies of 95.7\% $\pm$ 3.5\% and 96.1\% $\pm$ 1.5\%, respectively). These preliminary results are quite promising, given that the identification accuracy of human taxonomists are typically between the range of 67\% and 83\%, and thus illustrates the feasibility of leveraging machine learning and fluorescence-based spectral-morphological features as a viable method for automated identification of different algae types.
\end{abstract}

\section{Introduction} \label{sec:intro}

\begin{figure}[]
	\centering
	\includegraphics[width=\columnwidth]{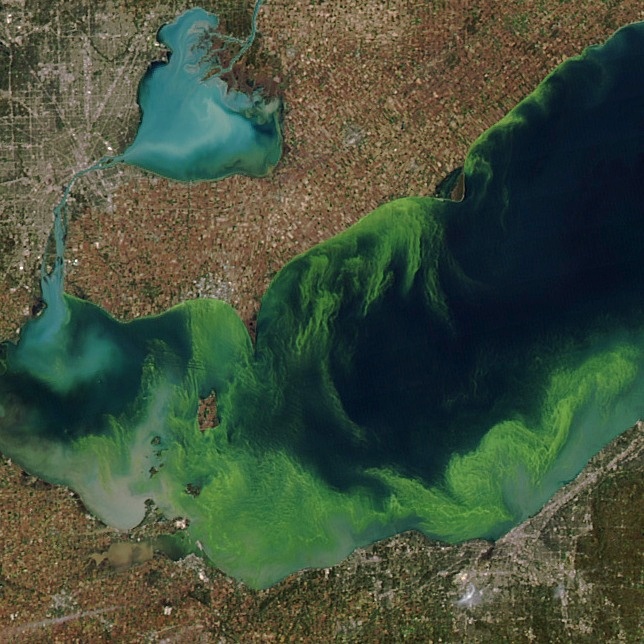}
	\caption{An Moderate Resolution Imaging Spectroradiometer (MODIS) image captured by the Aqua satellite showing Lake Erie on October 9, 2011.  The bloom was primarily \textit{Microcystis Aeruginosa}, according to the Great Lakes Environmental Research Laboratory, which is a common type of cyanobacteria~\cite{nasa_2011}.}
	\label{fig:lakeErie}
\end{figure}

Harmful algae blooms (HABs) are increasingly becoming a major threat to our water bodies.
An illustrative example of the threat of HABs is an incident in the summer of 2011, where Lake Erie experienced the largest harmful algae bloom in recorded history~\cite{michalak2013record} (see Figure~\ref{fig:lakeErie} for a Moderate Resolution Imaging Spectroradiometer (MODIS) image captured by the Aqua satellite of the incident).  This bloom was primarily \textit{Microcystis aeruginosa}, one of the most lethal cyanobacteria genera according to the Great Lakes Environmental Research Laboratory~\cite{nasa_2011}.

\begin{figure*}[]
	\centering
	\includegraphics[width=\textwidth]{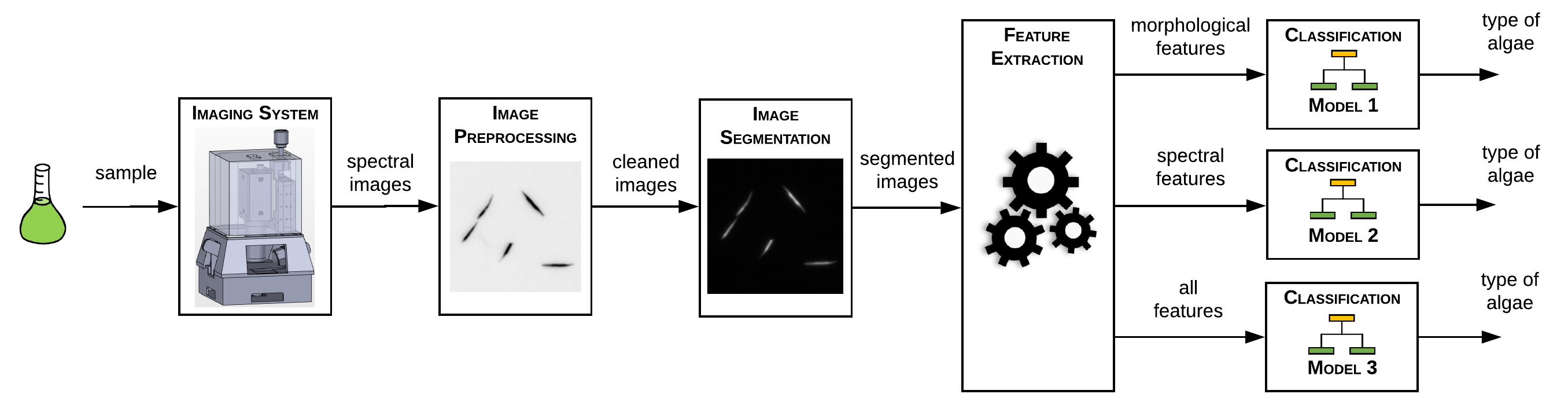}
	\caption{In this study, the proposed methodology for the purpose of automated identification of different algae types can be broken into five main steps.  First, a water sample containing algae was imaged using a custom multi-band fluorescence imaging microscope to capture fluorescence imaging data at a number of different excitation wavelengths (Section~\ref{imageAcq}).  The captured fluorescence imaging data is then processed using data processing algorithms for background subtraction (Section~\ref{preprocessing}).  Next, the processed fluorescence imaging data is first segmented into background objects and micro-organism objects, and the individual algae are isolated and extracted to produce isolated micro-organism imaging data (Section~\ref{segmenation}).  A number of morphological and spectral fluorescence features are then extracted from the isolated micro-organism imaging data, and used to train three different neural network models designed for the purpose of identification of algae types given an isolated micro-organism. (Section~\ref{featExtraction}).  More specifically, three different neural network models trained: i) using only the fluorescence-based morphological features (Model 1), ii) using only the fluorescence-based spectral features (Model 2), and iii) using fluorescence-based spectral-morphological features (Model 3).}
	\label{fig:flow}
\end{figure*}

For example, many types of cyanobacteria (a type of algae) can be extremely dangerous for humans and animals.  For example, swallowing \textit{Microcystis aeruginosa} can have serious side effects such as abdominal pain, diarrhea, vomiting, blistered mouths, dry coughs, and headaches~\cite{falconer1996potential}.  In addition, \textit{Anabaena sp.} can produce lethal neurotoxins called anatoxin-a which has been shown to cause death by progressive respiratory paralysis~\cite{falconer1996potential}.

In another illustrative example, a toxin produced by Microcystis, called Microcystin-LR (MC-LR), is strictly regulated by the World Health Organization (WHO) since it can be lethal for humans~\cite{world1998cyanobacterial}.
In addition, the maximum acceptable concentration (MAC) for the cyanobacterial toxin Microcystin-LR (MC-LR) in drinking water is 0.0015 mg/L, based on guidelines from the Government of Canada~\cite{2002HealthCanada}.
To better prevent toxin exposure during a bloom event, active monitoring of water quality is critical as it enables the collection of both temporal and spatial trends of bloom activity.  These spatial-temporal trends can then be inspected and analyzed via predictive analytic strategies, thus providing key stakeholders with early warning signs that a bloom may occur.

Commonly, the task of identification and enumeration of algae in natural water is conducted at certificated laboratory facilities, where highly-trained taxonomists perform manual analysis on the water samples~\cite{barsanti2014algae}.  This time-consuming process requires samples to be preserved, shipped, and then inspected at the laboratory facilities with expensive laboratory equipment.  As an alternative, in-situ monitoring devices such as fluorometric meters have been used to detect specific pigment levels, which can act as an indirect indicator for the purpose of monitoring in practice.  While this alternative approach can be useful for rough assessments of algae distributions, it still requires further manual confirmation due to the level of granularity in order to inform decision-makers to take the necessary actions needed to minimize exposure risk.  Furthermore, a study presented by Culverhouse \textit{et al.} show that human taxonomists have an identification accuracy between 67\% and 83\%, depending on the taxonomist~\cite{culverhouse2003experts}.  It is their conclusion is that the experts in the study are not unanimous in their identification, even when inspecting microorganisms with very distinct morphology~\cite{culverhouse2003experts, sieracki2010optical, colares2013microalgae, correa2016supervised}. Therefore, a method that could not only directly identity algae types in an automated and cost effective manner is highly desired for water industry.

In this study, we investigate the feasibility of leveraging machine learning and fluorescence-based spectral-morphological features to enable the identification of six different algae types in an automated fashion.  More specifically, we explore and investigate the efficacy of a number of different morphological and spectral fluorescence features extracted from multi-band fluorescence imaging data when used to train neural network classification models designed for the purpose of identification of six algae types in an automated manner.

The paper is organized as follows.  First, related work is discussed in Section~\ref{sec:relatedWork}. Second, the proposed methodology for investigating the feasibility of automated identification of algae types is presented in Section~\ref{sec:method}.  Third, the experimental setup used in this study is presented in Section~\ref{sec:results}.  Fourth, experimental results are presented and discussed in Section~\ref{sec:discussion}.  Finally, conclusions are drawn in Section~\ref{sec:conclusions}.

\section{Related Work} \label{sec:relatedWork}

A number of computational techniques and methods have been proposed for the purpose of automatically identification of different algae types using imaging data.  In the following section we present, in chronological order, a summary of the major contributions in the recent past in this area.

In 2002, Walker \textit{et al.} leveraged fluorescence excitation in their imaging protocol~\cite{walker2002fluorescence} for automated identification of \textit{Anabaena sp.} and \textit{Microcystis sp.}.  More specifically, their protocol was able to achieve over 97\% identification accuracy when looking for \textit{Anabaena sp.} and \textit{Microcystis sp.} in natural populations found in Lake Biwa, Japan, by capturing a single fluorescence image and a single brightfield image.
The authors claim that without the use of the fluorescence component, the automated identification of microalgae in the sediment saturated samples would be nearly impossible.

Similar results were found by  Hense \textit{et al.} in 2008 when they showed that by using epifluorescence microscopy in combination with brightfield microscopy they could correctly identify between 13 different phytoplankton samples as either algae or non-algae~\cite{hense2008use}.
They accomplished this by using three different filter sets to capture the fluorescence data and built a hand-tuned classifier based on empirically derived thresholds.
However, the main drawback of both these methods is they only tested classifiers with two classes, either \textit{Anabaena sp.} and \textit{Microcystis sp.} \cite{walker2002fluorescence} or algae or non-algae~\cite{hense2008use}.  Furthermore, both of these methods leveraged only a single fluorescence wavelength in combination with the brightfield image.

In 2010, Hu \textit{et al.} utilized the fact that different algae species have different ratios of antenna pigments, which results in different fluorescence emission spectra~\cite{hu2010multiple}.  More specifically, Hu \textit{et al.} illuminated twenty different algae from six algae divisions (Dinophyta, Bacillariophyta, Chrysophyta, Cyanophyta, Cryptophyta, and Chlorophyta) at four different wavelengths (440 nm, 470 nm, 530 nm, and 580 nm), and then measured the emission spectra from 600 nm - 750 nm with a 5 nm resolution.  By concatenating these four emission spectra together and conducting a multivariate linear regression and weighted least-squares it was found that each of the feature vectors from each phylum was independent from the others.  Although Hu \textit{et al.} showed that these feature vectors were independent, one major drawback of the method Hu \textit{et al.} proposed is that the relative ratios of different algae can only be achieved at the phyla level when mixing two species from different phyla.

More recently, Deglint~\textit{et al.} conducted a comprehensive spectral analysis of the fluorescence characteristics of three algae species when excited at twelve discrete spectral wavelengths~\cite{deglint2017comprehensive}.  Their findings was that the fluorescence spectra of the three algae species appear quite distinctive, and thus the use of multi-band fluorescence imaging microscopy could be a promising direction to explore.  However, that study is highly preliminary as the number of algae species studied was very limited and a more comprehensive quantitative investigation on how best to leverage such spectral information was not well explored for the purpose of automated identification of algae types.

Motivated by the findings of Deglint~\textit{et al.}~\cite{deglint2017comprehensive}, we aim to go a major step further by investigating and exploring the utilization of machine learning and fluorescence-based spectral-morphological features derived from multi-band fluorescence imaging microscopy data at different excitation wavelengths (between 405 nm - 530 nm) and a larger number of algae types (six different algae types in total).

\section{Methodology} \label{sec:method}

The proposed methodology used in this study to explore the feasibility of automated identification of different algae types using machine learning and fluorescence-based spectral-morphological features can be broken into five main steps (see Figure~\ref{fig:flow}).  First, a water sample containing algae was imaged using a custom multi-band fluorescence imaging microscope to capture fluorescence imaging data at a number of different excitation wavelengths (Section~\ref{imageAcq}).  The captured fluorescence imaging data are then processed using data processing algorithms for background subtraction (Section~\ref{preprocessing}).  Next, the processed fluorescence imaging data is first segmented into background objects and micro-organism objects, and the individual algae are isolated and extracted to produce isolated micro-organism imaging data (Section~\ref{segmenation}).  A number of morphological and spectral fluorescence features are then extracted from the isolated micro-organism imaging data, and used to train three different neural network models designed for the purpose of identification of algae types given an isolated micro-organism. (Section~\ref{featExtraction}).

\subsection{Data acquisition} \label{imageAcq}

\begin{figure}[t!]
	\centering
	\includegraphics[width=.88\columnwidth]{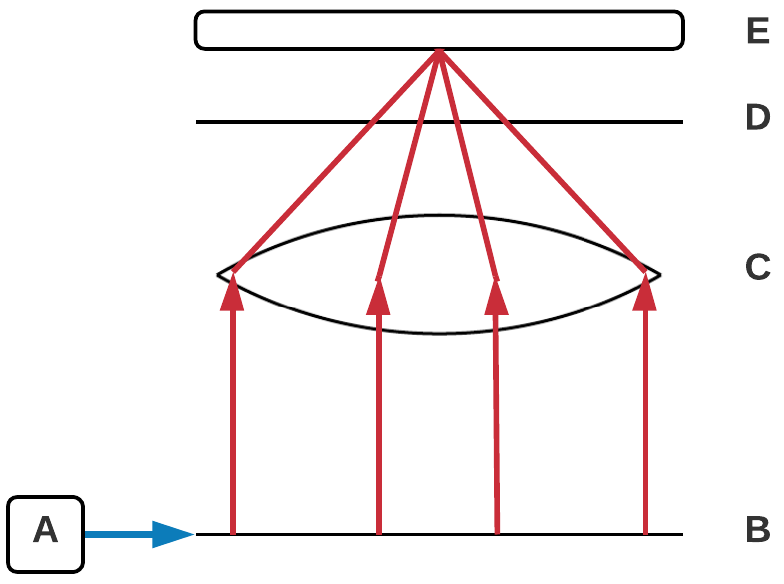}
	\caption{Image acquisition is performed using a custom multi-band fluorescence imaging microscope, which is composed of five main elements.  First, a light source at a particular excitation wavelength (A)  (blue arrows) is used to illuminate a water sample placed on a blank slide (B), thus effectively exciting the algae in the water sample.  The algae in the water sample then fluoresce and emit light at a lower energy (red arrows) and (C) the emitted light is focused using an magnification lens.  This focused light passes through (D) a highpass filter before hitting (E) a monochromatic sensor.}
	\label{fig:setup}
\end{figure}

The custom multi-band fluorescence imaging microscope used in this study for capturing fluorescence imaging data is composed of five main elements, as seen in Figure~\ref{fig:setup}.  First, a light source at a particular excitation wavelength (Figure~\ref{fig:setup}A)  (blue arrows) is used to illuminate a water sample placed on a blank slide (Figure~\ref{fig:setup}B), thus effectively exciting the algae in the water sample.  The algae in the water sample then fluoresce and emit light at a lower energy (red arrows) and (Figure~\ref{fig:setup}C) the emitted light is focused using an magnification lens.  This focused light passes through (Figure~\ref{fig:setup}D) a highpass filter before hitting (Figure~\ref{fig:setup}E) a monochromatic sensor.

Given that multiple wavelengths can be used to excite the algae, let $\lambda_1$, $\lambda_2$, to  $\lambda_m$ be the individual fluorescence images captured using different excitation wavelengths of a given water sample. Therefore, we define the entire multi-band fluorescence image $\Lambda_{raw}$ as
\begin{equation}
\Lambda_{raw} =
\begin{bmatrix}
\lambda_1 & \lambda_2 & \cdots & \lambda_i & \cdots & \lambda_m
\end{bmatrix}.
\label{eq:defineLambda}
\end{equation}

\subsection{Data processing} \label{preprocessing}

To improve the quality of the captured fluorescence imaging data obtained from the custom multi-band fluorescence imaging microscope for subsequent micro-organism isolation and neural network classification modeling steps, a set of automated data processing algorithms are first performed to compensate for some of the issues associated with the captured data.
First, to compensate for the fact that the illumination source in the imaging microscope does not illuminate the imaging field of view in a perfectly homogenous manner, iterative background subtraction is performed on each $\lambda_i$ in $\Lambda_{raw}$.  This background subtraction can be expressed as,

\begin{equation}
\Lambda_{corrected} =  \Lambda_{raw} - \Lambda_{background}
\end{equation}

\noindent where $\Lambda_{corrected}$ and $\Lambda_{background}$ denotes the illumination corrected imaging data and the background image, respectively.  To approximate the background image $\Lambda_{background}$, a Gaussian low-pass filter was first applied to $\Lambda_{raw}$ to perform noise suppression.  Next, to suppress features in $\Lambda_{raw}$ at different scales to better approximate $\Lambda_{background}$, an iterative multi-scale morphological opening was performed on the Gaussian low-pass filtered image, where the size of a disk structuring element is changed at each iteration.

\subsection{Image Segmentation and Micro-organism extraction} \label{segmenation}

Given the corrected multi-band fluorescence imaging data, $\Lambda_{corrected}$, we wish to now segment the background from the micro-organism objects, as well as isolate each micro-organism in the captured data.
This will allow features to be extracted from each segmented organism, which is vital when training and testing a given classification model.
To achieve segmentation of background from the micro-organism objects in the captured data, a binary background-foreground classifier was used to classify each pixel as either the foreground $C_f$ or the background $C_b$.
The advantage of using fluorescent images when segmenting is that there is a large contrast between the foreground and the background, allowing a single decision boundary to be found that can separate the organisms from the background.
To learn the decision boundary, $\theta$, of this binary background-foreground classifier, the within class variance was minimized, as originally proposed by Otsu \textit{et al.}~\cite{otsu1975threshold}.
Therefore, the binary background-foreground classifier can be expressed as,
\begin{equation}
L_i(\underline{x}) =
\begin{cases}
    C_f  & \text{if } f_i(\underline{x}) > \theta	\\
    C_b  & \text{otherwise}\\
  \end{cases}
\label{eq:binaryClassifier}
\end{equation}
\noindent where $f_i(\underline{x})$ is the pixel intensity at pixel $\underline{x}$ for a given wavelength image $\lambda_i$, where $i \in [1, m]$.

Given the segmented foreground-background information, a connected-connected analysis strategy was used to group neighbouring pixels in the foreground class together to isolate individual micro-organisms.  Each isolated micro-organism in the water sample can be defined as $\phi_j$, where $j \in [1, n]$, where $n$ is the total number of micro-organisms in the segmented image set $\Phi$.

\subsection{Fluorescence-driven Spectral-Morphological Feature Extraction and Neural Network Classification Modeling} \label{featExtraction}

\begin{center}
\begin{figure*}[t!]
	\centering
	\setlength{\tabcolsep}{3pt}
	\begin{tabular}{ccc}
		\includegraphics[width=.32\textwidth]{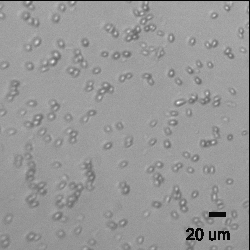}  &
		\includegraphics[width=.32\textwidth]{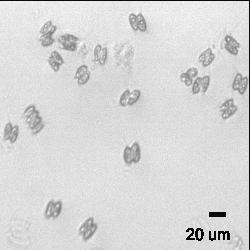}  &
		\includegraphics[width=.32\textwidth]{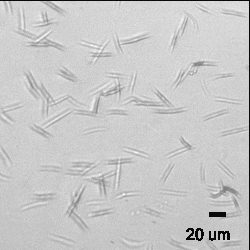}   \\
		\textbf{\textit{Scenedesmus obliquus}} & \textbf{\textit{Scenedesmus quadricauda}} & \textbf{\textit{Ankistrodesmus falcatus}} \\
		\textbf{Chlorophyta} & \textbf{Chlorophyta} & \textbf{Chlorophyta} \\		
		\textbf{CPCC 005} & \textbf{CPCC 158} & \textbf{CPCC 366} \\
						  & 				  & 				  \\
		\includegraphics[width=.32\textwidth]{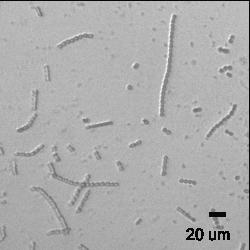}  &
		\includegraphics[width=.32\textwidth]{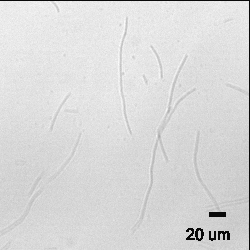}  &
		\includegraphics[width=.32\textwidth]{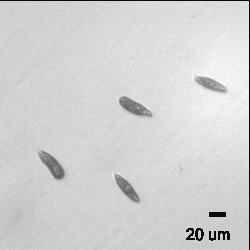}   \\
		\textbf{\textit{Anabaena flos-aquae}} & \textbf{\textit{Pseudanabaena tremula}} & \textbf{\textit{Euglena gracilis}} \\
		\textbf{Cyanophyta} & \textbf{Cyanophyta} & \textbf{Euglenozoa} \\		
		\textbf{CPCC 067} & \textbf{CPCC 471} & \textbf{CPCC 095} \\
	\end{tabular}	
	\caption{The six algae under investigation in this study are \textit{Scenedesmus obliquus} (CPCC 005), \textit{Scenedesmus quadricauda} (CPCC 158), \textit{Ankistrodesmus falcatus} (CPCC 366), \textit{Anabaena flos-aquae} (CPCC 067), \textit{Pseudanabaena tremula} (CPCC 471), and \textit{Euglena gracilis} (CPCC 095).  These algae are from three different phyla classes (Chlorophyta, Cyanophyta, and Euglenozoa) and two samples are from the same genus (\textit{Scenedesmus}).  Within the Cyanophyta phylum, two filamentous algae were chosen (\textit{Anabaena flos-aquae} (CPCC 067), \textit{Pseudanabaena tremula} (CPCC 471)) to determine how well a neural network classifier model driven by fluorescence-based spectral-morphological features can identify between two algae with very similar morphology.  These images are captured  using our custom system in brightfield mode, where the spatial resolution is 1.2 $\mu$m / pixel, as determined using the 1951 US Air Force (USAF) glass slide resolution target.}
	\label{fig:brightfieldImages}
\end{figure*}
\end{center}

Having isolated the micro-organisms in the imaging data, the goal is then to learn a classification model for identifying between the different types of algae in an automated manner. To achieve this, we first extract a set of fluorescence-based morphological and spectral features for characterizing individual micro-organisms.  The motivation for the proposed fluorescence-based morphological and spectral feature set is that, by leveraging not only features for characterizing the morphology of a micro-organism but also a greater number of spectral features gained from the custom-built fluorescence imaging microscope, a more complete profile can be constructed around the micro-organism to enable better discrimination between different types of algae than can be achieved using previous approaches.  In the proposed fluorescence-based spectral-morphological feature set, we leverage the five morphological features proposed in~\cite{gonzalez2012digital}, which can be described as follows:
\begin{enumerate}
\item \textbf{Area}: The total number of pixels in an isolated micro-organism.
\item \textbf{Convex Area}: The total number of pixels of the convex hull of an isolated micro-organism.
\item \textbf{Eccentricity}: The eccentricity of an isolated micro-organism is the ratio of the distance between the foci of the ellipse and its major axis, and is bounded between zero and one.
\item \textbf{Equivalent Diameter}: The diameter of a circle with the same area as an isolated micro-organism.
\item \textbf{Extent}: The ratio of pixels of an isolated micro-organism to the bounding box that contains that isolated micro-organism.
\end{enumerate}

Furthermore, we incorporate a set of spectral features for characterizing the mean of the fluorescence intensities in an isolated micro-organism, at each of the different captured excitation wavelengths.  In this study, since we capture at six different excitation wavelengths, the set of spectral features can be described as follows:

\begin{enumerate}
\item \textbf{Emission Signal 1}: The mean fluorescence intensity for an isolated micro-organism when excited at 405 nm.
\item \textbf{Emission Signal 2}: The mean fluorescence intensity for an isolated micro-organism when excited at 420 nm.
\item \textbf{Emission Signal 3}: The mean fluorescence intensity for an isolated micro-organism when excited at 450 nm.
\item \textbf{Emission Signal 4}: The mean fluorescence intensity for an isolated micro-organism when excited at 470 nm.
\item \textbf{Emission Signal 5}: The mean fluorescence intensity for an isolated micro-organism when excited at 500 nm.
\item \textbf{Emission Signal 6}: The mean fluorescence intensity for an isolated micro-organism when excited at 530 nm.
\end{enumerate}

Finally, given the fluorescence-based morphological and spectral feature vectors extracted from the isolated microorganisms, a classification model must be learned to predict the associated output class (algae type) given on these input feature vectors for the purpose of automated identification of algae types.  A number of machine learning approaches can be leveraged to learn the relationship between the input fluorescence-based morphological and spectral and the associated algae type, ranging from support vector machines~\cite{cortes1995support} to decision trees~\cite{breiman2017classification} and Naive Bayes~\cite{russell2016artificial}.

In this study, the classification models used are feedforward neural networks, which is an artificial neural network where information moves from the input layer, through a given amount of hidden layers, and then to the output layer.
An advantage in leveraging a feedforward neural network for the classification model is that a feedfoward neural network is an universal approximator~\cite{hornik1991approximation, lecun2015deep}.  As such, a feedforward neural network has the ability to approximate any continuous function with a finite number of neurons, and thus well-suited for learning a good approximation of a function that maps the fluorescence-based morphological and spectral features to the corresponding algae type.
Each layer of the network consists of multiple neurons that take a weighted sum of the inputs, $x_k$ and bias, $b$ and transform them with a non-linear activation function, $f(z)$.
This non-linear function takes as input
\begin{equation}
z = \sum_{k=1}^{n} x_k w_k + b
\end{equation}
\noindent where $w_k$ is a given weight for $n$ inputs~\cite{bishop2006pattern}.

Using these extracted morphological and spectral features in tandem with a feedforward neural network, three different neural network classification models are trained and evaluated.
\begin{itemize}
\item \textbf{Model 1}: Uses only the morphological features.
\item \textbf{Model 2}: Uses only the spectral features.
\item \textbf{Model 3}: Uses spectral-morphological features.
\end{itemize}

Model 1 is trained and tested using only the five extracted morphological features.  This model is used as a baseline to determine how well classification can be achieved when only looking at the size and shape characteristics of the algae.
Next, Model 2 is trained and tested using only the six extracted spectral features.  By only using these spectral features we can compare the relative performance of Model 1 and Model 2.
Finally, Model 3 is trained and tested using all the fluorescence-based spectral-morphological features.  This model will closely mimic what human taxonomists use when classifying different types of algae since it incorporates both the color as well as the shape and size of the algae.


\section{Experimental Setup} \label{sec:results}

To investigate and explore the feasibility of leveraging machine learning and fluorescence-based spectral-morphological features to enable the identification of different algae types in an automated fashion, a number of experiments were designed and tested.
First, we present the algae types selected for our experiments (Section~\ref{sec:algae}) and then discuss the hardware implementation to collect data (Section~\ref{sec:hardwareConfig}).
Finally, a full description of the three model architectures will be presented (Section~\ref{sec:modelArch}).

\subsection{Types of Algae} \label{sec:algae}

Six different types of algae from the Canadian
Phycological Culture Centre (CPCC) were chosen to explore the identification accuracy of the learned neural network models.  As seen in Figure~\ref{fig:brightfieldImages}, these six algae types, with their respective CPCC number, are broken into there respective phyla. These algae types along with the corresponding number of micro-organism samples for each type are described as follows:
\begin{enumerate}[I.]
\item{Chlorophyta (green algae)}
	\begin{enumerate}[{1.}]
	\item \textit{Scenedesmus obliquus} (CPCC 005): 751 samples
	\item \textit{Scenedesmus quadricauda} (CPCC 158): 382 samples
	\item \textit{Ankistrodesmus falcatus} (CPCC 366): 500 samples
	\end{enumerate}
\item{Cyanophyta (blue-green algae or cyanobacteria)}
	\begin{enumerate}[{1.}]
	\setcounter{enumii}{3}
	\item \textit{Anabaena flos-aquae} (CPCC 067): 548 samples
	\item \textit{Pseudanabaena tremula} (CPCC 471): 299 samples
	\end{enumerate}
\item{Euglenozoa}
	\begin{enumerate}[{1.}]
	\setcounter{enumii}{5}
	\item \textit{Euglena gracilis} (CPCC 095): 131 samples
	\end{enumerate}
\end{enumerate}

To build up this dataset, pure samples of each of the six types of algae were imaged with the custom-built fluorescence imaging microscope, which will be described in the next section.
A cropped section of the brightfield images can be seen in Figure~\ref{fig:brightfieldImages}.  This was captured by placing a white light source under the microscope slide and capturing an image with the custom-built fluorescence imaging microscope.

These samples were strategically chosen to be a broad representation of algae, given that three different phyla classes are present.  However, special attention was given to blue-green and green algae since they are the most common toxin producers in our waters.
Furthermore, in the Cyanophyta class, two very similar filamentous algae were chosen (\textit{Anabaena flos-aquae} and \textit{Pseudanabaena tremula}) to see how far our classification could differentiate between two similar filamentous algae.
Finally, two species from the \textit{Scenedesmus} genus were also chosen to see if classification down to the species level is possible.

\subsection{Hardware Configuration} \label{sec:hardwareConfig}

\begin{center}
\begin{figure*}[t!]
	\vspace{5mm}
	\centering
	\setlength{\tabcolsep}{3pt}
	\begin{tabular}{ccccccc}
		\raisebox{13 pt}[0 pt][0 pt]{\rotatebox{90}{\textbf{CPCC 005}}}	&
		\includegraphics[width=.15\textwidth]{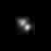}  &
		\includegraphics[width=.15\textwidth]{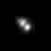}  &
		\includegraphics[width=.15\textwidth]{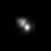}  &
		\includegraphics[width=.15\textwidth]{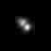}  &
		\includegraphics[width=.15\textwidth]{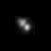}  &
		\includegraphics[width=.15\textwidth]{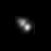}  \\
		\raisebox{13 pt}[0 pt][0 pt]{\rotatebox{90}{\textbf{CPCC 158}}}	&
		\includegraphics[width=.15\textwidth]{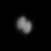}  &
		\includegraphics[width=.15\textwidth]{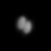}  &
		\includegraphics[width=.15\textwidth]{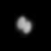}  &
		\includegraphics[width=.15\textwidth]{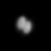}  &
		\includegraphics[width=.15\textwidth]{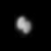}  &
		\includegraphics[width=.15\textwidth]{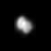}  \\
		\raisebox{13 pt}[0 pt][0 pt]{\rotatebox{90}{\textbf{CPCC 366}}}	&
		\includegraphics[width=.15\textwidth]{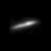}  &
		\includegraphics[width=.15\textwidth]{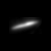}  &
		\includegraphics[width=.15\textwidth]{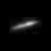}  &
		\includegraphics[width=.15\textwidth]{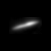}  &
		\includegraphics[width=.15\textwidth]{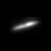}  &
		\includegraphics[width=.15\textwidth]{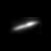}  \\
		\raisebox{13 pt}[0 pt][0 pt]{\rotatebox{90}{\textbf{CPCC 067}}}	&
		\includegraphics[width=.15\textwidth]{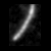}  &
		\includegraphics[width=.15\textwidth]{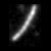}  &
		\includegraphics[width=.15\textwidth]{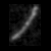}  &
		\includegraphics[width=.15\textwidth]{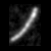}  &
		\includegraphics[width=.15\textwidth]{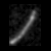}  &
		\includegraphics[width=.15\textwidth]{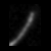}  \\
		\raisebox{13 pt}[0 pt][0 pt]{\rotatebox{90}{\textbf{CPCC 471}}}	&
		\includegraphics[width=.15\textwidth]{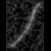}  &
		\includegraphics[width=.15\textwidth]{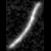}  &
		\includegraphics[width=.15\textwidth]{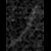}  &
		\includegraphics[width=.15\textwidth]{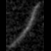}  &
		\includegraphics[width=.15\textwidth]{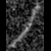}  &
		\includegraphics[width=.15\textwidth]{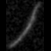}  \\
		\raisebox{13 pt}[0 pt][0 pt]{\rotatebox{90}{\textbf{CPCC 095}}}	&
		\includegraphics[width=.15\textwidth]{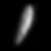}  &
		\includegraphics[width=.15\textwidth]{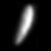}  &
		\includegraphics[width=.15\textwidth]{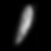}  &
		\includegraphics[width=.15\textwidth]{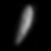}  &
		\includegraphics[width=.15\textwidth]{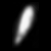}  &
		\includegraphics[width=.15\textwidth]{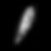}  \\
		 & \textbf{405 nm} & \textbf{420 nm} & \textbf{450 nm} & \textbf{470 nm} & \textbf{500 nm} & \textbf{530 nm}   \\
	\end{tabular}	
	\caption{Six types of algae from three phyla classes were imaged: \textit{Scenedesmus obliquus} (CPCC 005), \textit{Scenedesmus quadricauda} (CPCC 158), \textit{Ankistrodesmus falcatus} (CPCC 366), \textit{Anabaena flos-aquae} (CPCC 067), \textit{Pseudanabaena tremula} (CPCC 471), and \textit{Euglena gracilis} (CPCC 095).
Each algae was excited at six different wavelengths (405 nm, 420 nm, 450 nm, 470 nm, 500 nm, and 530 nm) and the fluorescent signal was captured with a monochrome sensor.
	Sample micro-organisms can be seen for each algae type at each excitation wavelength.
	The value of having this spectral information comes into play especially when the morphology between certain algae types are very similar in nature, such as in \textit{Anabaena flos-aquae} (CPCC 067) and \textit{Pseudanabaena tremula} (CPCC 471).}
	\label{fig:spectralImages}
\vspace{15mm}
\end{figure*}
\end{center}

\begin{figure}[t!]
	\centering
	\includegraphics[width=\columnwidth]{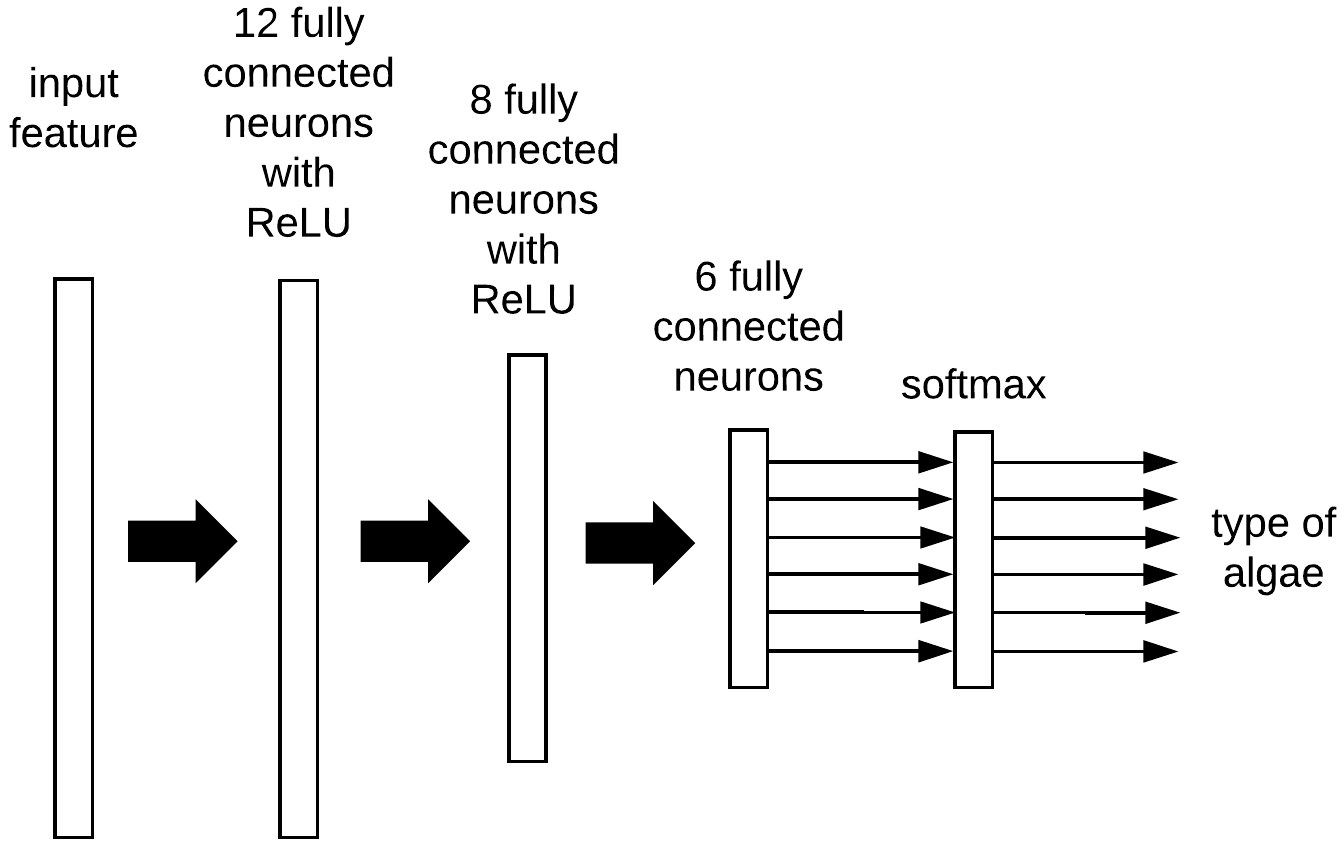}
	\caption{A fully-connected feed-forward neural network architecture was leveraged for all three neural network classification models.  The number of input features of this network is 5, 6, and 11, for Model 1, Model 2, and Model 3, respectively.  Model 1 only uses 5 morphological features, while Model 2 uses 6 spectral features.  Model three combines both these morphological and spectral features together to form a morphological-spectral feature set.}
	\label{fig:ANNarchitecture}
\end{figure}

The custom-built fluorescence imaging microscope contains six high power LEDs that emits light at six different spectral wavelengths (405 nm, 420 nm, 450 nm, 470 nm, 500 nm, and 530 nm).  These high-powered LEDs are placed orthogonal to a 3" x 1" microscope slide which had a pure algae sample on it as well as a standard cover slip.  The excited algae samples then emitted fluoresced light, which was focused by a 20x, passing through a 600 nm highpass filter and onto a 4.1 MP camera.
The spatial resolution of this system is 1.2 $\mu$m / pixel, as determined using the 1951 US Air Force (USAF) glass slide resolution target.
A sample image of each type of algae fluorescing at each excitation wavelength can be seen in Figure~\ref{fig:spectralImages}.

\subsection{Model Architectures} \label{sec:modelArch}

As previously discussed in Section~\ref{featExtraction}, three neural network classification models will be trained and tested to explore their relative identification performance.
The network architecture used for the three neural network classification models are based on a feedforward network architecture and can be seen in Figure~\ref{fig:ANNarchitecture}. The number of input features of this network is 5, 6, and 11, for Model 1, Model 2, and Model 3, respectively.  Model 1 only uses 5 morphological features, while Model 2 uses 6 spectral features.  Model three combines both these morphological and spectral features together to form a morphological-spectral feature set.
 The neural network classification models each contain three hidden layers, with the number of neurons at each layer being 12, 8, and 6, respectfully.
This decrease in the amount of neurons as the network continues to go deeper allows the input features to the transformed into a better representation for improve discrimination power between the different algae types.

In our network architecture design, we chose the rectified linear unit (ReLU) activation function for the  network, as defined as
\begin{equation}
H[i,j] = max(0, G[i,j]),
\end{equation}
\noindent where $H[i,j]$ is the output of the ReLU function.
Finally, the neural network classification models contain a softmax function, a normalized exponential function, which is used to transform the output to sum to one to mimic a probability mass function which can be defined as
\begin{equation}
\sigma(z^{(i)}) = \frac{e^{z^{(i)}}}{\sum_{k=1}^{n}e^{z^{(k)}}}
\end{equation}
\noindent where $k$ is the number of output classes, which is six in this study.

The neural network classification models were evaluated by using 20 runs of Monte Carlo Cross Validation (MCCV), where 70\% of the data was randomly selected
without replacement for a given run of the cross validation to be the training set.
The remaining 30\% of the input features were used as the test data-set.

\begin{table}[t!]
  \centering
  \caption{The mean and standard deviation of the identification accuracy using the test data across 20 test runs of Monte Carlo Cross Validation.  Model 1, trained using morphological features, had the lowest average accuracy and the highest standard deviation.  Model 2, trained using spectral features, and Model 3 trained using morphological-spectral features, achieved significantly higher average accuracies and lower standard deviations.}
    \begin{tabular}{|c|ccc|}
\cline{2-4}    \multicolumn{1}{c|}{} & \multicolumn{3}{c|}{\textbf{Accuracy}} \bigstrut\\
    \hline
    \textbf{Model 1} & 53.0\% & $\pm$     & 3.5\% \bigstrut\\
    \hline
    \textbf{Model 2} & 95.7\% & $\pm$     & 1.5\% \bigstrut\\
    \hline
    \textbf{Model 3} & 96.1\% & $\pm$     & 0.8\% \bigstrut\\
    \hline
    \end{tabular}%
  \label{tab:results}%
\end{table}%

\begin{table}[b!]
  \centering
  \caption{A pairwise t-test was run between all pairs of neural network classification models with a 1\% significance level.}
    \begin{tabular}{|c|c|c|}
    \hline
    \textbf{paired-sample t-test} & \textbf{Reject null hypothesis?} & \textbf{p-value} \bigstrut\\
    \hline
    Model 1 vs Model 2 & yes   & 7.35E-23 \bigstrut\\
    \hline
    Model 1 vs Model 3 & yes   & 6.04E-23 \bigstrut\\
    \hline
    Model 2 vs Model 3 & no    & 8.54E-01 \bigstrut\\
    \hline
    \end{tabular}%
  \label{tab:ttest}%
\end{table}%

\begin{table*}[]
	\vspace{3mm}
  \centering
  \caption{The best performing Model 1 from 20 Monte Carlo cross validation runs.}
    \begin{tabular}{|c|c|c|c|c|c|c|c|}
    \hline
    \multicolumn{2}{|c|}{\multirow{2}[4]{*}{\textbf{Model 1}}} & \multicolumn{6}{c|}{\textbf{PREDICTED}} \bigstrut\\
\cline{3-8}    \multicolumn{2}{|c|}{} & \textbf{CPCC 005} & \textbf{CPCC 158} & \textbf{CPCC 366} & \textbf{CPCC 067} & \textbf{CPCC 471} & \textbf{CPCC 095} \bigstrut\\
    \hline
    \multirow{6}[12]{*}{\textbf{TRUE}} & \textbf{CPCC 005} & 189   & 9     & \cellcolor[rgb]{ .855,  .588,  .58} \textbf{31} & 12    & 0     & 2 \bigstrut\\
\cline{2-8}          & \textbf{CPCC 158} & 4     & 70    & 8     & 7     & 0     & 10 \bigstrut\\
\cline{2-8}          & \textbf{CPCC 366} & 10    & 3     & 119   & 11    & 3     & 7 \bigstrut\\
\cline{2-8}          & \textbf{CPCC 067} & \cellcolor[rgb]{ .855,  .588,  .58} \textbf{65} & 10    & \cellcolor[rgb]{ .855,  .588,  .58} \textbf{50} & 21    & 11    & 9 \bigstrut\\
\cline{2-8}          & \textbf{CPCC 471} & 14    & 2     & \cellcolor[rgb]{ .855,  .588,  .58} \textbf{34} & 8     & 20    & 7 \bigstrut\\
\cline{2-8}          & \textbf{CPCC 095} & 2     & 6     & 5     & 2     & 1     & 21 \bigstrut\\
    \hline
    \end{tabular}%

  \label{tab:Model1}%
\end{table*}%

\begin{table*}[]
	\vspace{3mm}
   \centering
  \caption{The best performing Model 2 from 20 Monte Carlo cross validation runs.}
    \begin{tabular}{|c|c|c|c|c|c|c|c|}
    \hline
    \multicolumn{2}{|c|}{\multirow{2}[4]{*}{\textbf{Model 2}}} & \multicolumn{6}{c|}{\textbf{PREDICTED}} \bigstrut\\
\cline{3-8}    \multicolumn{2}{|c|}{} & \textbf{CPCC 005} & \textbf{CPCC 158} & \textbf{CPCC 366} & \textbf{CPCC 067} & \textbf{CPCC 471} & \textbf{CPCC 095} \bigstrut\\
    \hline
    \multirow{6}[12]{*}{\textbf{TRUE}} & \textbf{CPCC 005} & 220   & 0     & 0     & 0     & 0     & 0 \bigstrut\\
\cline{2-8}          & \textbf{CPCC 158} & 0     & 117   & 1     & 1     & 0     & \cellcolor[rgb]{ .855,  .588,  .58} \textbf{3} \bigstrut\\
\cline{2-8}          & \textbf{CPCC 366} & 0     & 0     & 160   & 0     & 0     & 0 \bigstrut\\
\cline{2-8}          & \textbf{CPCC 067} & 0     & 0     & 0     & 146   & \cellcolor[rgb]{ .855,  .588,  .58} \textbf{6}     & 0 \bigstrut\\
\cline{2-8}          & \textbf{CPCC 471} & 1     & 0     & 0     & 0     & 85    & 0 \bigstrut\\
\cline{2-8}          & \textbf{CPCC 095} & 0     & \cellcolor[rgb]{ .855,  .588,  .58} \textbf{4} & 1     & 0     & 1     & 37 \bigstrut\\
    \hline
    \end{tabular}%
  \label{tab:Model2}%
\end{table*}%

\begin{table*}[]
	\vspace{3mm}
  \centering
  \caption{The best performing Model 3 from 20 Monte Carlo cross validation runs.}
    \begin{tabular}{|c|c|c|c|c|c|c|c|}
    \hline
    \multicolumn{2}{|c|}{\multirow{2}[4]{*}{\textbf{Model 3}}} & \multicolumn{6}{c|}{\textbf{PREDICTED}} \bigstrut\\
\cline{3-8}    \multicolumn{2}{|c|}{} & \textbf{CPCC 005} & \textbf{CPCC 158} & \textbf{CPCC 366} & \textbf{CPCC 067} & \textbf{CPCC 471} & \textbf{CPCC 095} \bigstrut\\
    \hline
    \multirow{6}[12]{*}{\textbf{TRUE}} & \textbf{CPCC 005} & 205   & 0     & 0     & 0     & 0     & 0 \bigstrut\\
\cline{2-8}          & \textbf{CPCC 158} & 0     & 114   & 0     & 0     & 0     & 3 \bigstrut\\
\cline{2-8}          & \textbf{CPCC 366} & 0     & 0     & 165   & 0     & 0     & 0 \bigstrut\\
\cline{2-8}          & \textbf{CPCC 067} & \cellcolor[rgb]{ .855,  .588,  .58} \textbf{6} & 0     & 0     & 155   & 2     & 0 \bigstrut\\
\cline{2-8}          & \textbf{CPCC 471} & 0     & 0     & 0     & \cellcolor[rgb]{ .855,  .588,  .58} \textbf{5} & 81    & 0 \bigstrut\\
\cline{2-8}          & \textbf{CPCC 095} & 0     & 2     & 0     & 0     & 0     & 45 \bigstrut\\
    \hline
    \end{tabular}%
  \label{tab:Model3}%
\end{table*}%


\section{Experimental Results and Discussion} \label{sec:discussion}

The mean and standard deviation of identification accuracies across the 20 test runs of Monte Carlo Cross Validation can be seen in Table~\ref{tab:results}.  A number of observations can be made based on the identification accuracy results.  First of all, it was observed that Model 1, which utilizes only the set of five morphological features had the lowest average identification accuracy at 53.0\%.  Model 2, which leverages the set of six fluorescence-based spectral features significantly outperformed Model 1, with an average identification accuracy of 95.7\%.  Finally, Model 3, which leverages the combined fluorescence-based morphological-spectral feature set also demonstrate a strong average identification accuracy at 96.1\%.  Therefore, it can be shown that the utilization of fluorescence-based spectral features is very important for the automated identification of different algae types, and further extends upon the observations made by Walker et al.~\cite{walker2002fluorescence} regarding the necessity of spectral information.  It can also be observed that the standard deviation is significantly lower for Model 2 compared to Model 1, which indicates that the utilization of spectral features also provide more consistent identification performance across different permutations of samples, which is important for generalizability in real-world scenarios.  In addition, it can be observed that, when comparing Model 2 and Model 3, the average identification accuracy for Model 3 is increased compared to Model 2, as well as had a lower standard deviation.  What this means is that by also leveraging morphological features, Model 3 can provide consistently improved identification performance.

Since the increase in average identification accuracy in Model 3 compared to Model 2 was relatively low, a pairwise t-test was run between all pairs of two models with a 1\% significance level, as seen in Table~\ref{tab:ttest}.  As expected, the pairwise t-test for Model 1 vs Model 2 as well as Model 1 vs Model 3 show that the improvements of Model 2 over Model 1, as well as the improvements of Model 3 over Model are both statistically significant in terms of identification accuracy.
However, it was also found that Model 2 and Model 3 are do not show a statistically significant difference in their identification accuracy.
Therefore, based on these results, it was observed that the set of fluorescence-based spectral features leveraged in the study is the major contributor to improved identification accuracy performance.

Now, comparing the preliminary results achieved using the neural network classification models to that achieved by human taxonomists, which have an identification accuracy between 67\% and 83\% \cite{culverhouse2003experts}, it can be observed that Model 1 dramatically under-performed compared to human taxonomists, demonstrating that the utilization of just morphological features is not a reliable way to identify between different types of algae via imaging.
However, both Model 2 and Model 3 demonstrated very encouraging identification accuracies when compared to that of human taxonomists, and thus illustrate not only that fluorescence-based spectral features are highly effective for identifying between different types of algae, but also that such automated identification methods can be a very valuable tool for human taxonomists to leverage to reduce the time-consuming and tedious task of manually isolating and analyzing individual micro-organisms, and thus be able to spend more time on the more important judgement and assessment of water quality and determining the appropriate course of action to take to mitigate the situation.

\begin{center}
\begin{figure*}[t!]
	\vspace{3mm}
	\centering
	\setlength{\tabcolsep}{3pt}
        \begin{tabular}{ccccc}
    \textbf{Model 1}	 &
    \makecell{\includegraphics[width=.18\textwidth]{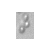}} &
    \makecell{\includegraphics[width=.18\textwidth]{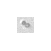}} &
    \makecell{\includegraphics[width=.18\textwidth]{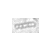}} &
    \makecell{\includegraphics[width=.18\textwidth]{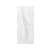}} \\
    \hline
    TRUE  &
    \textit{\makecell{Scenedesmus \\obliquus \\CPCC 005}} &
    \textit{\makecell{Anabaena \\flos-aquae \\CPCC 067}} &
    \textit{\makecell{Anabaena \\flos-aquae \\CPCC 067}} &
    \textit{\makecell{Pseudanabaena \\tremula \\CPCC 471}} \\
    \hline
    PREDICTED &
    \textit{\makecell{Ankistrodesmus \\falcatus \\CPCC 366}} &
    \textit{\makecell{Scenedesmus \\obliquus \\CPCC 005}} &
    \textit{\makecell{Ankistrodesmus \\falcatus \\CPCC 366}} &
    \textit{\makecell{Ankistrodesmus \\falcatus \\CPCC 366}} \\
    \hline
    \end{tabular}%
	\caption{Misidentified samples from Model 1 corresponding to errors highlighted in red from Table~\ref{tab:Model1}.}
	\label{fig:Model1_wrong}
\end{figure*}
\end{center}

\begin{center}
\begin{figure*}[t!]
	\vspace{3mm}
	\centering
	\setlength{\tabcolsep}{3pt}
        \begin{tabular}{ccccc}
    \textbf{Model 2}	 &
    \makecell{\includegraphics[width=.18\textwidth]{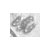}} &
    \makecell{\includegraphics[width=.18\textwidth]{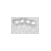}} &
    \makecell{\includegraphics[width=.18\textwidth]{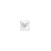}} &
    \makecell{\includegraphics[width=.18\textwidth]{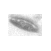}} \\
    \hline
    TRUE  &
    \textit{\makecell{Scenedesmus \\quadricauda \\CPCC 158}} &
    \textit{\makecell{Anabaena \\flos-aquae \\CPCC 067}} &
    \textit{\makecell{Anabaena \\flos-aquae \\CPCC 067}} &
    \textit{\makecell{Euglena \\gracilis \\CPCC 095}} \\
    \hline
    PREDICTED &
    \textit{\makecell{Euglena \\gracilis \\CPCC 095}} &
    \textit{\makecell{Pseudanabaena \\tremula \\CPCC 471}} &
    \textit{\makecell{Pseudanabaena \\tremula \\CPCC 471}} &
    \textit{\makecell{Scenedesmus \\quadricauda \\CPCC 158}} \\
    \hline
    \end{tabular}%
	\caption{Misidentified samples from Model 2 corresponding to errors highlighted in red from Table~\ref{tab:Model2}.}
	\label{fig:Model2_wrong}
\end{figure*}
\end{center}

\begin{center}
\begin{figure*}[t!]
	\vspace{3mm}
	\centering
	\setlength{\tabcolsep}{3pt}
        \begin{tabular}{ccccc}
    \textbf{Model 3}	 &
    \makecell{\includegraphics[width=.18\textwidth]{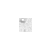}} &
    \makecell{\includegraphics[width=.18\textwidth]{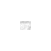}} &
    \makecell{\includegraphics[width=.18\textwidth]{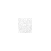}} &
    \makecell{\includegraphics[width=.18\textwidth]{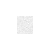}} \\
    \hline
    TRUE  &
    \textit{\makecell{Anabaena \\flos-aquae \\CPCC 067}} &
    \textit{\makecell{Anabaena \\flos-aquae \\CPCC 067}} &
    \textit{\makecell{Pseudanabaena \\tremula \\CPCC 471}} &
    \textit{\makecell{Pseudanabaena \\tremula \\CPCC 471}} \\
    \hline
    PREDICTED &
    \textit{\makecell{Scenedesmus \\obliquus \\CPCC 005}} &
    \textit{\makecell{Scenedesmus \\obliquus \\CPCC 005}} &
    \textit{\makecell{Anabaena \\flos-aquae \\CPCC 067}} &
    \textit{\makecell{Anabaena \\flos-aquae \\CPCC 067}} \\
    \hline
    \end{tabular}%
	\caption{Misidentified samples from Model 3 corresponding to errors highlighted in red from Table~\ref{tab:Model3}.}
	\label{fig:Model3_wrong}
\end{figure*}
\end{center}

\vspace{-2.5cm}
To gain a deeper understanding of the identification error associated with each neural network classification model, the confusion matrix of the best performing run from each 20 Monte Carlo Cross Validation runs for Model 1, Model 2, and Model 3 can be seen in Table~\ref{tab:Model1}, Table~\ref{tab:Model2}, and Table~\ref{tab:Model3}, respectively.  In these tables, the major sources of identification error have been highlighted in red.  Furthermore, selected miss-classified samples from each of these highlighted red errors can be seen in Figure~\ref{fig:Model1_wrong}, Figure~\ref{fig:Model2_wrong}, and Figure~\ref{fig:Model3_wrong}, for Model 1, Model 2, and Model 3, respectively.

In Table~\ref{tab:Model1} and Figure~\ref{fig:Model1_wrong}, it can be observed that for Model 1, which uses only morphological features, the greatest source of misidentifications came from \textit{Anabaena flos-aquae} (CPCC 067) and \textit{Pseudanabaena tremula} (CPCC 471) samples being mislabeled as \textit{Ankistrodesmus falcatus} (CPCC 366).  By inspecting both Figure~\ref{fig:brightfieldImages} and Figure~\ref{fig:spectralImages} it is clear that both of these species can have a elongated shape, which can cause problems with identification when leveraging only morphological features.
The other most common misidentification error came from \textit{Anabaena flos-aquae} (CPCC 067) being mislabeled as \textit{Scenedesmus obliquus} (CPCC 005).  Once again by looking at Figure~\ref{fig:brightfieldImages} and Figure~\ref{fig:spectralImages} smaller \textit{Anabaena flos-aquae} (CPCC 067) could easily be confused with an \textit{Scenedesmus obliquus} (CPCC 005) sample using just morphological features.

When inspecting Table~\ref{tab:Model2} and Figure~\ref{fig:Model2_wrong}, it can be observed that for Model 2, which uses only spectral features, the misidentification error was significantly less compared to that of Model 1, with the largest source of error being 4 samples of \textit{Euglena gracilis} (CPCC 095) being mislabeled as \textit{Scenedesmus quadricauda} (CPCC 158) and 3 samples of \textit{Scenedesmus quadricauda} (CPCC 158) were mislabeled as \textit{Euglena gracilis} (CPCC 095).  These results are consistent with the observation that the spectral features between \textit{Scenedesmus quadricauda} (CPCC 158) and \textit{Euglena gracilis} (CPCC 095) is much closer compared to other samples.    The other main source of error is when \textit{Anabaena flos-aquae} (CPCC 067) is classified as \textit{Pseudanabaena tremula} (CPCC 471).  This miss-classification is intuitive as both of these species are both from the Cyanophyta phylum and are both filamentous types of algae.

Finally, in Table~\ref{tab:Model3} and Figure~\ref{fig:Model2_wrong}, it can be observed that for Model 3, which utilized fluorescence-based morphological-spectral features, the misidentification error was significantly less compared to that of Model 1, with sources of error being 6 samples of CPCC 067 being mislabeled as \textit{Scenedesmus obliquus} (CPCC 005) and five samples of \textit{Pseudanabaena tremula} (CPCC 471) being mislabeled as \textit{Anabaena flos-aquae} (CPCC 067). This particular source of misidentification may be attributed to the fact that both are filamentous types with very similar shapes and characteristics.
In addition, as seen in Figure~\ref{fig:Model2_wrong}, certain \textit{Anabaena flos-aquae} (CPCC 067) were classified as \textit{Scenedesmus quadricauda} (CPCC 158) when the chain structure of the \textit{Anabaena flos-aquae} broke apart, leaving one or two single-celled organisms on their own.  When inspecting Figure~\ref{fig:brightfieldImages}, a given \textit{Anabaena flos-aquae} can easily be mistaken as a \textit{Scenedesmus quadricauda}.
Furthermore, when inspecting Figure~\ref{fig:Model2_wrong}, it is observed that some images, such as \textit{Pseudanabaena tremula} (CPCC 471) have not been well segmented.  This reveals that the classification error is primarily due to errors in the segmentation.

Therefore, our approach of capturing data at multiple wavelengths and processing it with a neural network shows promise that on-site monitoring of algae types is potentially possible.
Given that the identification accuracy of human taxonomists are typically between the range of 67\% and 83\%, our method illustrates the feasibility of leveraging machine learning and fluorescence-based spectral-morphological features as a viable method for automated identification of different algae types.
Future work involves the design of more advanced neural network classification architectures for handling scenarios characterized by the need to automatically identify a greater number of algae types in contaminated and mixed water samples.
Such a system will allow for near real-time analysis of a water sample to determine which types of algae are present as well as their relative concentrations.
This will give water treatment plants and other organizations the ability to build up a database of algae activity over time allowing them an early warning sign that bloom might occur.

\section{Conclusions} \label{sec:conclusions}

In this paper, the feasibility of leveraging machine learning and fluorescence-based spectral-morphological features for automated identification of algae type was explored.  In particular, neural network classification models were trained to identify different algae types using fluorescence-based spectral features and morphological features extracted from imaging data captured using a custom multi-band fluorescence imaging microscope at six different excitation wavelengths (405 nm, 420 nm, 450 nm, 470 nm, 500 nm, and 530 nm).  Experimental results using three different neural network models (one trained on morphological features, one trained on spectral fluorescence features, and one trained on spectral-morphological fluorescence features) on six different algae types (\textit{Scenedesmus obliquus} (CPCC 005), \textit{Scenedesmus quadricauda} (CPCC 158), \textit{Ankistrodesmus falcatus} (CPCC 366), \textit{Anabaena flos-aquae} (CPCC 067), \textit{Pseudanabaena tremula} (CPCC 471), and \textit{Euglena gracilis} (CPCC 095)) demonstrated that neural network classification models trained using either fluorescence-based spectral features or fluorescence-based spectral-morphological features resulted in average identification accuracies of 95.7\% and 96.1\%, respectively.  As such, the results of this study illustrate that leveraging machine learning and fluorescence-based spectral-morphological features can be a feasible direction for further exploration for the purpose of automated identification of different algae types.

\section*{Contributions}

JLD, CJ, and AW designed the functionality of the hardware system.  AC and JLD designed the form factor of the hardware system.  JLD collected the data used in the experiments. JLD and AW designed the data pipeline and architecture of the different machine learning classification models.  JLD, CJ, and AW conducted the analysis and contributed to the writing of the manuscript.

\bibliographystyle{IEEEtran}
\bibliography{refs}

\begin{thebibliography}{10}
\providecommand{\url}[1]{#1}
\csname url@samestyle\endcsname
\providecommand{\newblock}{\relax}
\providecommand{\bibinfo}[2]{#2}
\providecommand{\BIBentrySTDinterwordspacing}{\spaceskip=0pt\relax}
\providecommand{\BIBentryALTinterwordstretchfactor}{4}
\providecommand{\BIBentryALTinterwordspacing}{\spaceskip=\fontdimen2\font plus
\BIBentryALTinterwordstretchfactor\fontdimen3\font minus
  \fontdimen4\font\relax}
\providecommand{\BIBforeignlanguage}[2]{{%
\expandafter\ifx\csname l@#1\endcsname\relax
\typeout{** WARNING: IEEEtran.bst: No hyphenation pattern has been}%
\typeout{** loaded for the language `#1'. Using the pattern for}%
\typeout{** the default language instead.}%
\else
\language=\csname l@#1\endcsname
\fi
#2}}
\providecommand{\BIBdecl}{\relax}
\BIBdecl

\bibitem{michalak2013record}
A.~M. Michalak, E.~J. Anderson, D.~Beletsky, S.~Boland, N.~S. Bosch, T.~B.
  Bridgeman, J.~D. Chaffin, K.~Cho, R.~Confesor, I.~Dalo{\u{g}}lu
  \emph{et~al.}, ``Record-setting algal bloom in lake erie caused by
  agricultural and meteorological trends consistent with expected future
  conditions,'' \emph{Proceedings of the National Academy of Sciences}, vol.
  110, no.~16, pp. 6448--6452, 2013.

\bibitem{nasa_2011}
\BIBentryALTinterwordspacing
``Toxic algae bloom in lake erie,'' \emph{NASA}, Oct 2011. [Online]. Available:
  \url{https://earthobservatory.nasa.gov/IOTD/view.php?id=76127}
\BIBentrySTDinterwordspacing

\bibitem{falconer1996potential}
I.~R. Falconer, ``Potential impact on human health of toxic cyanobacteria,''
  \emph{Phycologia}, vol.~35, no.~6S, pp. 6--11, 1996.

\bibitem{world1998cyanobacterial}
W.~H. Organization \emph{et~al.}, ``Cyanobacterial toxins: microcystin-lr,''
  \emph{Guidelines for drinking water quality}, vol.~2, 1998.

\bibitem{2002HealthCanada}
H.~Canada, ``Canadian drinking water guidelines,'' \emph{Cyanobacterial Toxins
  – Microcystin–LR}, July 2002.

\bibitem{barsanti2014algae}
L.~Barsanti and P.~Gualtieri, \emph{Algae: anatomy, biochemistry, and
  biotechnology}.\hskip 1em plus 0.5em minus 0.4em\relax CRC press, 2014.

\bibitem{culverhouse2003experts}
P.~F. Culverhouse, R.~Williams, B.~Reguera, V.~Herry, and S.~Gonz{\'a}lez-Gil,
  ``Do experts make mistakes? a comparison of human and machine identification
  of dinoflagellates,'' \emph{Marine Ecology Progress Series}, vol. 247, pp.
  17--25, 2003.

\bibitem{sieracki2010optical}
M.~E. Sieracki, M.~Benfield, A.~Hanson, C.~Davis, C.~H. Pilskaln, D.~Checkley,
  H.~M. Sosik, C.~Ashjian, P.~Culverhouse, R.~Cowen \emph{et~al.}, ``Optical
  plankton imaging and analysis systems for ocean observation,''
  \emph{Proceedings of ocean Obs}, vol.~9, pp. 21--25, 2010.

\bibitem{colares2013microalgae}
R.~G. Colares, P.~Machado, M.~de~Faria, A.~Detoni, V.~Tavano \emph{et~al.},
  ``Microalgae classification using semi-supervised and active learning based
  on gaussian mixture models,'' \emph{Journal of the Brazilian Computer
  Society}, vol.~19, no.~4, pp. 411--422, 2013.

\bibitem{correa2016supervised}
I.~Corr{\^e}a, P.~Drews, M.~S. de~Souza, and V.~M. Tavano, ``Supervised
  microalgae classification in imbalanced dataset,'' in \emph{Intelligent
  Systems (BRACIS), 2016 5th Brazilian Conference on}.\hskip 1em plus 0.5em
  minus 0.4em\relax IEEE, 2016, pp. 49--54.

\bibitem{walker2002fluorescence}
R.~F. Walker, K.~Ishikawa, and M.~Kumagai, ``Fluorescence-assisted image
  analysis of freshwater microalgae,'' \emph{Journal of microbiological
  methods}, vol.~51, no.~2, pp. 149--162, 2002.

\bibitem{hense2008use}
B.~A. Hense, P.~Gais, U.~J{\"u}tting, H.~Scherb, and K.~Rodenacker, ``Use of
  fluorescence information for automated phytoplankton investigation by image
  analysis,'' \emph{Journal of Plankton Research}, vol.~30, no.~5, pp.
  587--606, 2008.

\bibitem{hu2010multiple}
X.~Hu, R.~Su, F.~Zhang, X.~Wang, H.~Wang, and Z.~Zheng, ``Multiple excitation
  wavelength fluorescence emission spectra technique for discrimination of
  phytoplankton,'' \emph{Journal of Ocean University of China}, vol.~9, no.~1,
  pp. 16--24, 2010.

\bibitem{deglint2017comprehensive}
J.~L. Deglint, J.~Chao, and A.~Wong, ``A comprehensive spectral analysis of the
  auto-fluorescence characteristics of three algae species at twelve discrete
  excitation wavelengths,'' \emph{Journal of Computational Vision and Imaging
  Systems}, vol.~3, no.~1, 2017.

\bibitem{otsu1975threshold}
N.~Otsu, ``A threshold selection method from gray-level histograms,''
  \emph{Automatica}, vol.~11, no. 285-296, pp. 23--27, 1975.

\bibitem{gonzalez2012digital}
R.~C. Gonzalez and R.~E. Woods, ``Digital image processing,'' 2012.

\bibitem{cortes1995support}
C.~Cortes and V.~Vapnik, ``Support-vector networks,'' \emph{Machine learning},
  vol.~20, no.~3, pp. 273--297, 1995.

\bibitem{breiman2017classification}
L.~Breiman, \emph{Classification and regression trees}.\hskip 1em plus 0.5em
  minus 0.4em\relax Routledge, 2017.

\bibitem{russell2016artificial}
S.~J. Russell and P.~Norvig, \emph{Artificial intelligence: a modern
  approach}.\hskip 1em plus 0.5em minus 0.4em\relax Malaysia; Pearson Education
  Limited,, 2016.

\bibitem{hornik1991approximation}
K.~Hornik, ``Approximation capabilities of multilayer feedforward networks,''
  \emph{Neural networks}, vol.~4, no.~2, pp. 251--257, 1991.

\bibitem{lecun2015deep}
Y.~LeCun, Y.~Bengio, and G.~Hinton, ``Deep learning,'' \emph{nature}, vol. 521,
  no. 7553, p. 436, 2015.

\bibitem{bishop2006pattern}
C.~M. Bishop, \emph{Pattern recognition and machine learning}.\hskip 1em plus
  0.5em minus 0.4em\relax springer, 2006.

\end{thebibliography}






\end{document}